\documentclass{article}
\usepackage{spconf,amsmath,graphicx}
\usepackage{url}

\title{Data Driven Grapheme-to-Phoneme Representations for a Lexicon-Free\\ Text-to-Speech }
\name{
Abhinav Garg$^{*}$\thanks{$^{*}$ Equal contributions. $^{\dagger}$Work done at Samsung Research.}$^{1\dagger}$, Jiyeon Kim$^{*2}$, Sushil Khyalia$^{3\dagger}$, Chanwoo Kim$^{4\dagger}$, Dhananjaya Gowda$^2$}

\address{
\texttt{abhinavg$@$stanford$.$edu}, \texttt{jstacey$7.$kim$@$samsung$.$com},
\texttt{skhyalia$@$andrew$.$cmu$.$edu},\\
\texttt{chanwcom$@$korea$.$ac$.$kr},
\texttt{d$.$gowda$@$samsung$.$com}\\
\vspace{-10pt}$^1$Stanford University$\qquad$ $^2$Samsung Research$\qquad$ $^3$Carnegie Mellon University$\qquad$ $^4$Korea University }

\vspace{-20pt}
\begin{document}
%
\maketitle
\begin{abstract}
Grapheme-to-Phoneme (G2P) is an essential first step in any modern, high-quality Text-to-Speech (TTS) system. Most of the current G2P systems rely on carefully hand-crafted lexicons developed by experts. This poses a two-fold problem. Firstly, the lexicons are generated using a fixed phoneme set, usually, ARPABET or IPA, which might not be the most optimal way to represent phonemes for all languages. Secondly, the man-hours required to produce such an expert lexicon are very high. In this paper, we eliminate both of these issues by using recent advances in self-supervised learning to obtain data-driven phoneme representations instead of fixed representations. 
We compare our lexicon-free approach against strong baselines that utilize a well-crafted lexicon.
Furthermore, we show that our data-driven lexicon-free method performs as good or even marginally better than the conventional rule-based or lexicon-based neural G2Ps in terms of Mean Opinion Score (MOS) while using no prior language lexicon or phoneme set, i.e. no linguistic expertise.
\end{abstract}
\section{Introduction}
\label{sec:intro}

Text-to-Speech synthesis has been the subject of extensive research for several decades \cite{  8461368, DBLP:conf/ssw/OordDZSVGKSK16}. 
Initially, concatenative speech synthesis models  were developed to address this task by assembling waveforms from a pre-existing database of speech. Subsequently, statistical approaches were introduced to generate speech features from the text, which were then fed to a vocoder to produce the final output. The results of these methods are unsatisfactory due to the unnaturalness and mispronunciations in the generated speech.

Grapheme-to-Phoneme (G2P) models are an integral part of current Text-to-Speech (TTS) engines~\cite{neuralG2P, neural_speech_synthesis}.
The initial work in this field was done using rule-based and joint sequence models. However, with the rise of deep learning methods, RNN, and even more recently, Transformer based architectures have been used to perform a variety of tasks such as Grapheme-to-Phoneme, Automatic Speech Recognition\cite{garg2019improved, Garg2020StreamingOE, Garg2020HierarchicalMW}, Machine Translation, and Text-to-Speech Synthesis  \cite{neuralG2P, neural_speech_synthesis,  8462506, JiyeonASRU2021}.
As G2P conversion is essentially a sequence-to-sequence modeling task, using Encoder-Decoder architectures\cite{Kim2021Steamingseven, Gowda2019MultiTaskMC} helped obtain improvements on Grapheme-to-Phoneme conversion \cite{neuralG2P}.

Traditional G2Ps \cite{cmudict} typically use a large lexicon to perform dictionary searches of the most frequent words and use hand-crafted rules to generate pronunciations for out-of-vocabulary words. Neural G2Ps~\cite{neuralG2P}, in contrast, use lexicons as their data for training their neural network and use the obtained network for predicting pronunciations. While both of these methods have been instrumental in building modern-day TTS systems, they have severe limitations in terms of requiring an external lexicon. Building a lexicon is an expensive and highly cumbersome task, as it requires multiple language experts to propose and then verify its validity.

 Recently there have been efforts to build a massively multi-lingual ByT5 G2P~\cite{Zhu2022ByT5MF}. 
 It uses a T5 transformer based encoder-decoder architecture~\cite{Rezckov2021T5G2PUT} and uses an UTF8 based input tokenization to handle scripts in multiple languages.
 It uses publicly available lexicons from the internet covering upto 100 languages with varying size and quality of the lexicons.
 The difficulty in procuring high quality lexicon for all languages and the decreasing accuracy of g2p models for languages with limited or noisy data can be clearly seen from this paper.

In this paper, we propose a new mechanism to train the Grapheme-to-Phoneme model without the need of any lexicon. We use Text-to-Speech as a use-case to show the effectiveness of our G2P model. We first use unlabeled speech data to pre-train a HuBERT \cite{9585401} for three iterations. Once we have a pre-trained HuBERT model, we input the labeled speech data and use a  specific transformer layer to extract the speech features and apply k-means clustering on them to obtain frame-level phoneme targets. We use paired phoneme targets and labeled speech data to train our G2P transformer model. We then use the trained G2P to train a Tacotron 2 \cite{8461368} model.


\section{Neural Grapheme-to-Phoneme}
\label{sec:format}

\label{sec:neural_g2p}
\subsection{G2P Architecture}
Transformer-based Grapheme-to-Phoneme models have been found to outperform LSTM / Recurrent Encoder-Decoder models  \cite{neuralG2P, 7178767, yao2015sequence} while also having the advantage of faster parallel training resulting in significantly less training time.

The neural Grapheme-to-Phoneme model takes the grapheme sequence $X = (x_1,x_2, \dots, x_n)$ as input. It then generates representations $L = (l_1, l_2, \dots, l_n)$ in a latent space which are then passed to a decoder. Finally, the decoder tries to predict the corresponding phoneme sequence $Y = (y_1, y_2, \dots, y_m)$ in an autoregressive manner. Here $n, m$ is the token length of input and output respectively.


\subsection{Text-to-Speech System}
To evaluate our G2P for the TTS task, we use Tacotron 2 \cite{8461368} as our text-to-speech model. Although the original Tacotron 2 uses text, many studies \cite{valle2021flowtron, https://doi.org/10.48550/arxiv.2205.04421} have found phones to produce superior results. Hence, we  use phones as inputs to the Tacotron model for our experiments. 

\begin{figure*}
    \centering
    \includegraphics[width=15cm]{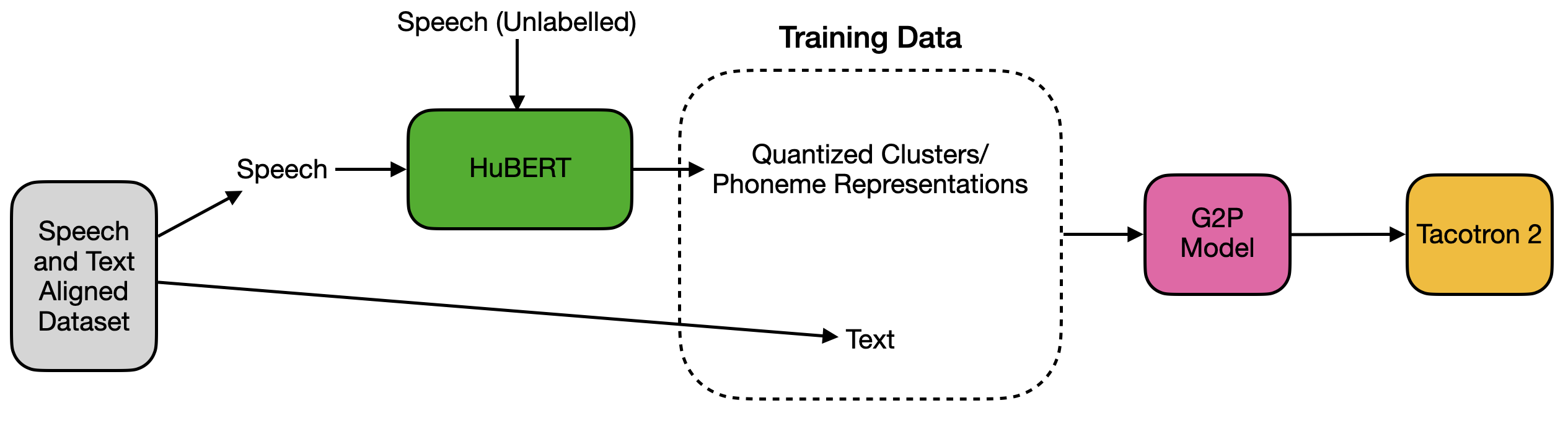}
    \caption{Flow chart demonstrating proposed algorithm}
    \label{fig:diagram_flow}
\end{figure*}
\section{Self Supervision in Speech Recognition}
\label{sec:pagestyle}

Models like wav2vec2.0 \cite{NEURIPS2020_92d1e1eb} quantize and mask input audio signals, after which they use contrastive loss to allow the model to learn from the unlabeled data. In contrast, models like HuBERT \cite{9585401} separate the acoustic unit discovery step from the masked prediction representation learning stage, allowing for a more direct predictive loss computation. We use these acoustic units in Section \ref{sec:algo} for training our G2P models. We outline the pre-training approach for obtaining these acoustic units in HuBERT below.

Given speech representation feature sequence $X = [x_1, x_2, ..., x_T]$ of length $T$, $M \subset [T]$, a corrupted sequence $\tilde X $ is obtained by replacing input representation at time $t$, $x_t$ with $\tilde x$ if $t \in M$, where $\tilde x$ is a learned vector. HuBERT encoder $e$ takes $\tilde X$ as input and predicts an output distribution over target vocabulary at each time step $p(z_t|\tilde X, t)$, where $z_t$ is frame level target obtained via acoustic unit discovery. 

For obtaining frame-level targets, an acoustic unit discovery model is used. Given $X$, acoustic unit discovery model $h$ takes $X$ as input to produce $h(X) = Z = [z_1, z_2, ... , z_T]$ where each $z_t$ belongs to a $C$-class categorical distribution and $h$ is a clustering model. We use the same clustering model as \cite{9585401} i.e., k-means.

While training the model, $M$ is calculated by randomly selecting $p\%$ of timesteps as the starting indices and masking $l$ steps starting from them. Cross entropy loss is then calculated over the masked timesteps.
\begin{equation}
    L_m(e;X,M,Z) = \Sigma_{t \in M} log\ p(z_t|\tilde X, t), 
\end{equation}

Encoder $e$ is trained to minimize $L_m$ in the pre-training stage.
The pre-training is done in multiple iterations. In the initial iteration, the clustering model $h$ uses MFCC-generated features for producing $Z$. However, as the training proceeds, to obtain better targets $Z$, $h$ uses latent features extracted from the HuBERT model pre-trained in the previous iteration at some intermediate layer.
\section{Data Driven Grapheme-to-Phoneme for Text-to-Speech}
\label{sec:algo}

Recent TTS systems \cite{8461368, Wang2017TacotronTE} have shown remarkable results in producing almost natural-sounding audios. However, these TTS systems require equally high-performing G2P systems to produce these high-quality audios. Building a well-performing G2P poses a couple of challenges for both traditional lexicon based G2Ps and neural G2Ps. For traditional G2P systems, obtaining a lexicon is a tedious task and requires experts. Moreover, even if the lexicon can be obtained, making rules for out-of-vocabulary words poses another issue. Although neural G2P does not use a lexicon for inference, it uses a lexicon as training data.

While obtaining a lexicon for a language is laborious, speech data is generally available in abundance. Obtaining a text transcription corresponding to a few hours of speech data is much more feasible. Starting with unlabeled speech data and a small amount of labeled data, we propose a novel training strategy to (1) Obtain a data-driven phoneme set in contrast to the traditional sets like CMU (ARPABET) or IPA, (2) Train a neural G2P on this new phoneme set, (3) We subsequently use the trained G2P for training a TTS system.

For achieving this, we use the acoustic units produced by a HuBERT model during the discovery stage as our phonetic representation. Using the number of clusters (i.e., $k$-value), we can control the size of the phoneme set. We present our training algorithm with the availability of unlabeled speech dataset $S_u$  and a small labeled dataset $S_l = [(s_1,t_1),...(s_n,t_n)]$, where $(s_i,t_i)$ is a single data point with $s_i$ audio utterance paired with $t_i$ text. A flow chart of the algorithm is shown in Fig. \ref{fig:diagram_flow} and explained below.

\begin{enumerate}
    \item We use the unlabeled speech data $S_u$ to pre-train a HuBERT for three iterations. Using MFCC for the first iteration, then using some intermediate transformer layer. See Section \ref{sec:exp}. 
    \item Once we have a pre-trained HuBERT model, we input the labeled speech data, $(s_1, ... , s_n)$, and use the 9th transformer layer to extract the speech features and apply k-means clustering on them to obtain frame-level phoneme targets $(P_1,...., P_n)$, where each $P_i = h(q_i) = [z^i_1, ... z^i_T]$ and $q_i$ is the 9th layer output of $e(s_i)$.
    \item Once we have $P_i$ for all $s_i$ in $S_l$, we use the paired $(P_i, t_i)$ to train our G2P transformer model as described in Section \ref{sec:neural_g2p}.
    \item We then use the trained G2P model to train a Tacotron 2 model described in Section \ref{sec:neural_g2p}.
\end{enumerate}

Using the above training algorithm, we can leverage a large amount of available unlabeled speech data and train high-quality TTS models with a minimal amount of labeled training data.

\section{Experiments}
\label{sec:exp}
\subsection{HuBERT model}
\label{subsec:exp:HuBERT}
As a first step, We trained the HuBERT Base model~\cite{9585401} with 95M parameters. 39 dimension MFCC features were used for obtaining cluster targets in the initial iteration, after which the 9th transformer layer outputs were used. A total of 3 iterations were performed, and the 9th transformer layer outputs of the final iteration were also used in obtaining final phoneme targets to train the G2P model. 

We used the LibriSpeech dataset to pre-train the model. 
We only use the 960h training set for pre-training HuBERT model and no fine-tuning stage was performed. 

For masking, we randomly mask $8\%$ of the timesteps with a mask length of 10 steps. An initial value of k=100 for k-means clustering was used in the first iteration, after which k=500 was used. However, while obtaining phone targets, k=100 was used. Adam optimizer with linear decay and peak learning rate of $5e^{-4}$ was used. A learning rate warmup was also used for the first $8\%$ of the training steps.
\subsection{G2P Data}
To obtain training data for our G2P model, we used the publically available LJSpeech Dataset \cite{ljspeech17} with $\sim$ 24hrs of single speaker data at a sampling frequency of 22050. We split the dataset into standard training, validation, and testing splits of 8:1:1, respectively. Audio data from LJSpeech was downsampled to 16000 before passing it through HuBERT. And as mentioned in Section \ref{subsec:exp:HuBERT}, we used k=100 for k-means clustering to obtain phone targets.

To train our rule-based and neural baselines mentioned in Section \ref{sec:neural_g2p}, we use the CMU dict dataset. The training, validation, and testing splits were the same as 
\cite{neuralG2P} to make the results directly comparable.

\subsection{G2P model}
For the G2P transformer model, we used a modified version of the open-source toolkit Deep Phonemizer\footnote{\url{https://github.com/as-ideas/DeepPhonemizer}} \cite{Yolchuyeva_2019}.
We use 512 as the size for transformer input features. Both the encoder and decoder have 4 layers of transformers. We use 1024 as the size for the feed-forward network. 
A dropout rate of 0.1 was used along with a multi-head attention consisting of 4 heads. We used a learning rate of 0.0001 with the Adam optimizer. The input text was chunked into words before passing them through the network.

\subsection{Tacotron 2}
We adopted the open-source TTS library CoquiTTS\footnote{ \url{https://github.com/coqui-ai/TTS}} for building and training our TTS models. 
Original sampling rate data of 22050 Hz was used without any downsampling. Adam optimizer was used with $\beta_1 = 0.9, \beta_2 = 0.998, \epsilon = 10^{-6}$ and a learning rate of $1e^{-4}$. Wav portions smaller than 40dB were considered silent and were eliminated from the dataset. 

\subsection{Evaluation}
We calculate the usability of our models and compare our model with baselines using the golden mean opinion score (GMOS) and mean opinion score (MOS) respectively. For calculating GMOS and MOS, we synthesized 40
sentences from the test set. Each sentence was rated by 10 Human evaluators on a 10 points Likert scale from 0.5 (“Bad”) to 5 (“Excellent”) with a step size of 0.5. We report the results with 95\% confidence interval.

\begin{table}[]
\caption{\label{tab:cmu-dict}Comparison of our transformer G2P model with other previous works.}

\begin{tabular}{|c|c|c|}
\hline
Model & PER\% & WER\% \\ \hline
    Bi-directional LSTM (3 layers) \cite{yao2015sequence}  &    5.45  & 23.55    \\
    Encoder CNN, decoder Bi-LSTM \cite{app9061143}	&4.81	&25.13      \\
    Ours &  5.25   &   26.34 \\ \hline
\end{tabular}

\end{table}

\section{Results}
\label{sec:res}

To verify the quality of our G2P transformer model, we first train our model and obtain the results on the standard CMU dict dataset. As can be seen from Table \ref{tab:cmu-dict}, our models are comparable to the previous state-of-the-art. We used the ARPABET phoneme set as the output phoneme set for training. Phoneme Error Rate (PER) is obtained by calculating the Levenshtein distance between the predicted phoneme sequence and the reference phoneme sequence in CMU dict. In case of multiple pronunciations, the one with the minimum distance is used. Word Error Rate (WER) measures how many word phoneme sequences were predicted exactly by the transformer model. 

To evaluate the usability of our approach, we calculate the GMOS using the HuBERT model.
In this setup, we don't use any G2P model instead use the HuBERT model directly to produce phoneme sequences using the test audio. All the HuBERT configurations were the same as those mentioned earlier. As we use phonemes obtained from HuBERT model to train our G2P model, these phoneme sequences are like golden targets for us, and their good quality would indicate the usefulness of exploring the usage of HuBERT outputs to train G2P systems. Using test audios to produce phonemes, we obtained a MOS of \textbf{$4.2 \pm 0.06$}, which is significantly better than the baseline, demonstrating the usage of HuBERT outputs as phonemes for TTS systems\footnote{Audio samples are available at : \url{https://abhinavg4.github.io/g2p/}}.

Furthermore, we compare our G2P trained model using a custom phoneme set produced by the algorithm outlined in Section \ref{sec:algo} with traditional rule-based G2P~\cite{Phonetisaurus} and neural G2P trained using a lexicon. The results shown in Table \ref{tab:g2p-mos} clearly indicate that our algorithm produces superior results as compared to the traditional approach or even the neural approach. Apart from producing better results, our approach does not need a lexicon and hence can be used more widely.

\subsection{Ablation Studies}
Finally, we perform some ablation studies by varying the number of values of clusters ($k$) when producing phoneme targets for the G2P model. Intuitively a large value of $k$ would allow a wider range of phonemes to be produced, which in turn would allow for better audio production in TTS. However, a very large value of $k$ would add little to the variety and would make the model bulkier. From Table \ref{tab:cluster} we observe that the value of 100 works best in the case of the English language. While the value of 50 is too small to represent the variety of sounds present, the value of 150 is too high and leads to worse performance. 

\begin{table}[]
\caption{\label{tab:g2p-mos}MOS results for comparison between our proposed neural G2P, traditional baseline and neural baseline.} 
\begin{tabular}{|c|c|}
\hline
            Model                  & Mean Opinion Score (MOS) \\ \hline
Rules based G2P               &         3.8 $\pm$ 0.04                \\
Neural G2P using Lexicon &              3.91    $\pm$ 0.06        \\ 
Proposed Neural G2P           &         \textbf{3.96 $\pm$ 0.05}                 \\ 
Natural Speech & 4.32 $\pm$ 0.07 \\ \hline
Golden Hubert Labels & 4.2 $\pm$ 0.06 \\ \hline

\end{tabular}

\end{table}

\begin{table}[]
\caption{\label{tab:cluster}Ablation results for impact of cluster size on MOS.}
\centering
\begin{tabular}{|c|c|c|}
\hline
k & Mean Opinion Score (MOS) \\ \hline
50 & 3.71 $\pm$ 0.08 \\ 
100 & 3.96 $\pm$ 0.06 \\ 
150 & 3.82 $\pm$ 0.06 \\ \hline
\end{tabular}

\end{table}

\section{Conclusion }
In this work, we explore a data-driven way to obtain phoneme representation for a language. We then used these phoneme representations to train a neural Grapheme-to-Phoneme model and used that model to train a high-quality TTS system. Apart from performing comparable to previous G2Ps, our G2P does not require any domain experts to build the lexicon. We perform our experiments on the English language to compare our results with previous works. Finally, we present ablation studies on the size of phoneme set optimal for the English Language. As our methods does not require linguist expertise our method can be easily extended for low-resource languages as a future work.
\bibliographystyle{IEEEbib}
\bibliography{strings, refs, mybib}

\end{document}